\documentclass[english]{lni}
\usepackage{subfig}
\newcommand{\PaperAcronym}{EdgeMLOps\xspace}

\begin{document}
\title[Operationalizing ML models]{\PaperAcronym: Operationalizing ML models with Cumulocity IoT and thin-edge.io for Visual quality Inspection}
 \author[1]{Kanishk Chaturvedi}{Kanishk.Chaturvedi@Cumulocity.com}{}
 \author[2]{Johannes Gasthuber}{Johannes.Gasthuber@Siemens.com}{}
 \author[3]{Mohamed Abdelaal}{Mohamed.Abdelaal@Softwareag.com}{}
 \affil[1]{Cumulocity GmbH\\Toulouser Allee 25\\ 40211 Düsseldorf\\ Germany}
 \affil[2]{Siemens AG\\Hertha-Sponer-Weg 3\\ 91058 Erlangen\\ Germany}
 \affil[3]{Software AG\\Uhlandstrasse 12\\64297 Darmstadt\\Germany}
\maketitle

\begin{abstract}
This paper introduces \PaperAcronym, a framework leveraging Cumulocity IoT and thin-edge.io for deploying and managing machine learning models on resource-constrained edge devices. We address the challenges of model optimization, deployment, and lifecycle management in edge environments. The framework's efficacy is demonstrated through a visual quality inspection (VQI) use case where images of assets are processed on edge devices, enabling real-time condition updates within an asset management system. Furthermore, we evaluate the performance benefits of different quantization methods, specifically static and dynamic signed-int8, on a Raspberry Pi 4, demonstrating significant inference time reductions compared to FP32 precision. Our results highlight the potential of \PaperAcronym to enable efficient and scalable AI deployments at the edge for industrial applications.
\end{abstract}
\begin{keywords}
MLOps, Edge AI, Cumulocity IoT, thin-edge.io, ML Inference 
\end{keywords}
\vspace{-3mm}\section{Introduction}\label{sec:intro}

Machine learning (ML) inference at the edge, encompassing on-device computation for tasks like image recognition, natural language processing, and sensor data analysis, offers compelling advantages over cloud-based inference. These include reduced latency, enhanced privacy, and increased availability, particularly crucial for real-time applications and scenarios with limited or intermittent connectivity~\cite{10637271}. Edge inference empowers applications ranging from autonomous vehicles and smart home devices to industrial automation and remote healthcare monitoring, fostering greater responsiveness and efficiency. However, deploying sophisticated ML models on resource-constrained edge devices presents significant challenges. Edge devices face challenges due to limited computational power, memory, and energy compared to cloud systems. To deploy complex models effectively, they must be optimized and compressed, while also accommodating the diverse architectures and conditions of edge environments.


To overcome these challenges, we introduce \PaperAcronym{}, an architectural framework for operationalizing ML models at the edge. The framework includes two main components, including Cumulocity IoT\footnote{\url{https://cumulocity.com/}} and the open-source thin-edge.io tool\footnote{\url{https://thin-edge.io/}}. Cumulocity IoT is a robust, cloud-native platform designed to facilitate the management and integration of connected devices and applications. It offers a comprehensive suite of tools for device management, data visualization, and real-time analytics, enabling businesses to rapidly develop and deploy IoT solutions while ensuring scalability, security, and ease of use. Thin-edge.io is an open-source, cloud-agnostic framework specifically designed for resource-constrained IoT edge devices, enabling efficient deployment and management of lightweight IoT agents. 

The integration of Cumulocity IoT and thin-edge.io effectively addresses the challenges associated with deploying machine learning models at the edge. First, by leveraging thin-edge.io's lightweight and modular architecture, organizations can optimize and compress complex ML models to fit within the limited computational power and memory capacity of edge devices without significantly sacrificing accuracy. Second, Cumulocity IoT enhances device management through features such as over-the-air (OTA) updates and centralized monitoring, which are essential for maintaining model performance in dynamic edge environments. The platform's ability to manage diverse hardware and operating systems allows for adaptable deployment strategies that cater to the heterogeneity of edge devices. Additionally, the integration facilitates robust on-device adaptation and update mechanisms, enabling seamless adjustments to fluctuating network conditions and evolving data distributions

To show the effectiveness of \PaperAcronym{}, we use the framework to address the pains in a use case of visual quality inspection. In this use case, field engineers use a mobile app to capture images of hardware assets (like transformers and switches) during inspections. These images are processed locally using an AI-powered Visual Quality Inspection (VQI) module. The VQI module uses a trained ML model to identify the asset type and its health status. This information is continuously updated in the asset management system, allowing managers to optimize maintenance and replacement schedules. In this use case, several challenges often arise that \PaperAcronym{} aims to address. Updating a fleet of edge devices with the latest model release can be cumbersome, and rolling back to earlier versions in response to detected production issues adds another layer of complexity. Additionally, adapting models for heterogeneous devices and new deployment environments, particularly those with lower-end hardware, can be a costly engineering effort. Collecting and labeling fresh data samples from the deployment environment requires substantial supervision and effort, especially when deploying to independent edge devices. \PaperAcronym{} is designed to streamline these processes, reducing the associated burdens and costs.

Several frameworks have been developed to support edge AI by enabling efficient deployment and management of ML models on resource-constrained devices. Tools like Google’s Edge TPU\footnote{https://cloud.google.com/edge-tpu} and NVIDIA’s Jetson platform\footnote{https://developer.nvidia.com/embedded-computing} focus on hardware acceleration, offering high-performance inferencing capabilities. Open-source frameworks like Apache NiFi\footnote{https://nifi.apache.org/} and EdgeX Foundry\footnote{https://www.edgexfoundry.org/} provide modular architectures for edge data processing and analytics but lack native support for ML model lifecycle management. EdgeMLOps distinguishes itself by integrating Cumulocity IoT for centralized device management and thin-edge.io for lightweight, cloud-agnostic deployment, addressing the unique requirements of industrial IoT environments. Furthermore, the integration of ONNX Runtime enhances the portability of ML models across diverse hardware platforms, making EdgeMLOps adaptable and scalable.

To summarize, this paper offers several key contributions: (1) an architectural framework for deploying ML models on edge devices, (2) the application of this framework to a real-world use case in VQI-based asset management within industrial environments, and (3) a comparison of different quantization methods for compressing ML models and performing inference on a Raspberry Pi 4 device. The paper is organized as follows: Section~\ref{sec:case_Study} provides an in-depth look at the VQI use case.
Section~\ref{sec:pre} introduces later used fundamental tools and technologies.
Section~\ref{sec:architecture} introduces \PaperAcronym{} and its components. Section~\ref{sec:evaluation} compares quantization methods based on average inference time. Finally, Section~\ref{sec:conclusion} concludes with insights on the presented ideas.

%
\vspace{-3mm}\section{VQI Use Case}\label{sec:case_Study}

Power transmission poles are deployable field assets that eventually need renewal. A typical application tracking these assets might already be in place, but field engineers must evaluate the assets periodically. While one might be able to predict life expectancy with historical data, a failure in power transmission has both economic and human impacts and, therefore, needs precise insurance. A more dynamic approach with visual quality inspection could leverage AI models to detect the assets and classify their current quality, thereby enabling a higher and cheaper evaluation frequency. Taking the images and evaluating them could be done by less skilled professionals or even an autonomous vehicle or drone. Especially if only low cellular bandwidth is available, an edge deployment, either to the autonomous field engineer or an edge-based server, might be necessary to give early feedback on the quality of the pictures and alarm corresponding service technicians. 

\begin{figure}
    \centering
    \includegraphics[width=1\linewidth]{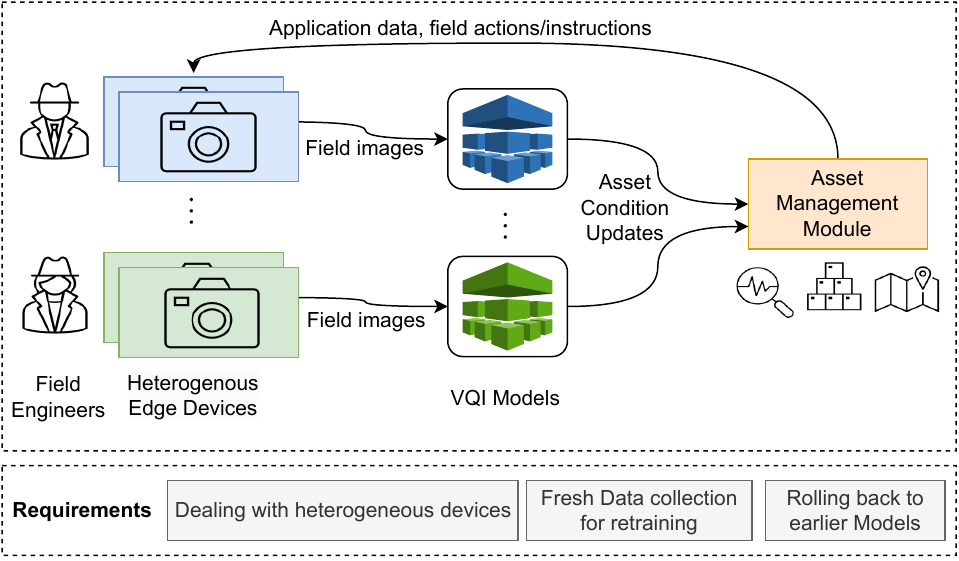}
    \caption{VQI use case and its requirements}
    \label{fig:general-architecture}
\end{figure}

Figure~\ref{fig:general-architecture} illustrates a workflow of the VQI use case involving field engineers and heterogeneous edge devices. Field engineers use these devices to capture images of assets, which are then processed by VQI models (e.g., ResNet101 implemented in PyTorch). The processed data generates asset condition updates that are sent to an asset management module. Such a module utilizes this information to conduct frequent and on-demand analyses, enhancing decision-making processes for maintenance and operational efficiency. The system also supports the collection of fresh data for model retraining and allows for rollbacks to previous models if needed. This setup addresses key requirements such as handling diverse devices and ensuring model adaptability in dynamic environments.


As the paper focuses on the IoT architecture and the corresponding technologies involved in preparing and deploying VQI models, the open-source TTPLA dataset \cite{abdelfattah2020ttpla}
has been utilized to train a ResNet50 and ResNet101 segmentation model for inference~\cite{clement2024xai}. 
This enables the free sharing of data and models to the community. TTPLA is a public dataset which is a collection of aerial images of transmission towers (TTs) and power lines (PLs). TTPLA supports the evaluation of instance segmentation, besides detection and semantic segmentation. 

\vspace{-3mm}\section{Preliminaries}\label{sec:pre}
%
In this section, we provide an overview of the tools and technologies used in \PaperAcronym to facilitate efficient inference at the edge. Specifically, we discuss the capabilities of Cumulocity IoT, thin-edge.io, and the ONNX standard. Cumulocity IoT is a comprehensive cloud platform designed to simplify the complexities of managing and operating Internet of Things (IoT) devices at scale. It supports a wide range of functionalities, including \textit{Device Connectivity}, which ensures seamless integration and communication across various IoT devices. \textit{Streaming Analytics} enables real-time data processing and analysis, allowing users to make immediate decisions based on live data streams. The platform also provides \textit{Application Enablement}, which facilitates the creation and deployment of custom IoT applications tailored to specific business needs. Central to its capabilities is robust \textit{Device Management}, which enables users to oversee, control, and secure a vast array of IoT devices from a centralized interface. 

Another critical feature of Cumulocity is its \textit{Software Repository}, a versatile component that allows users to upload and manage additional software artifacts. This repository is pivotal for deploying updates and new functionalities across connected devices efficiently, ensuring that the latest software is always at the forefront of operations. Furthermore, Cumulocity IoT provides robust security measures, including device authentication, encryption, and access control, addressing the critical need for data protection in edge deployments. The platform also offers advanced diagnostics and troubleshooting tools, allowing administrators to quickly identify and resolve issues with edge devices, minimizing downtime and ensuring consistent ML model performance. 

\begin{figure}
    \centering
    \includegraphics[width=1\linewidth]{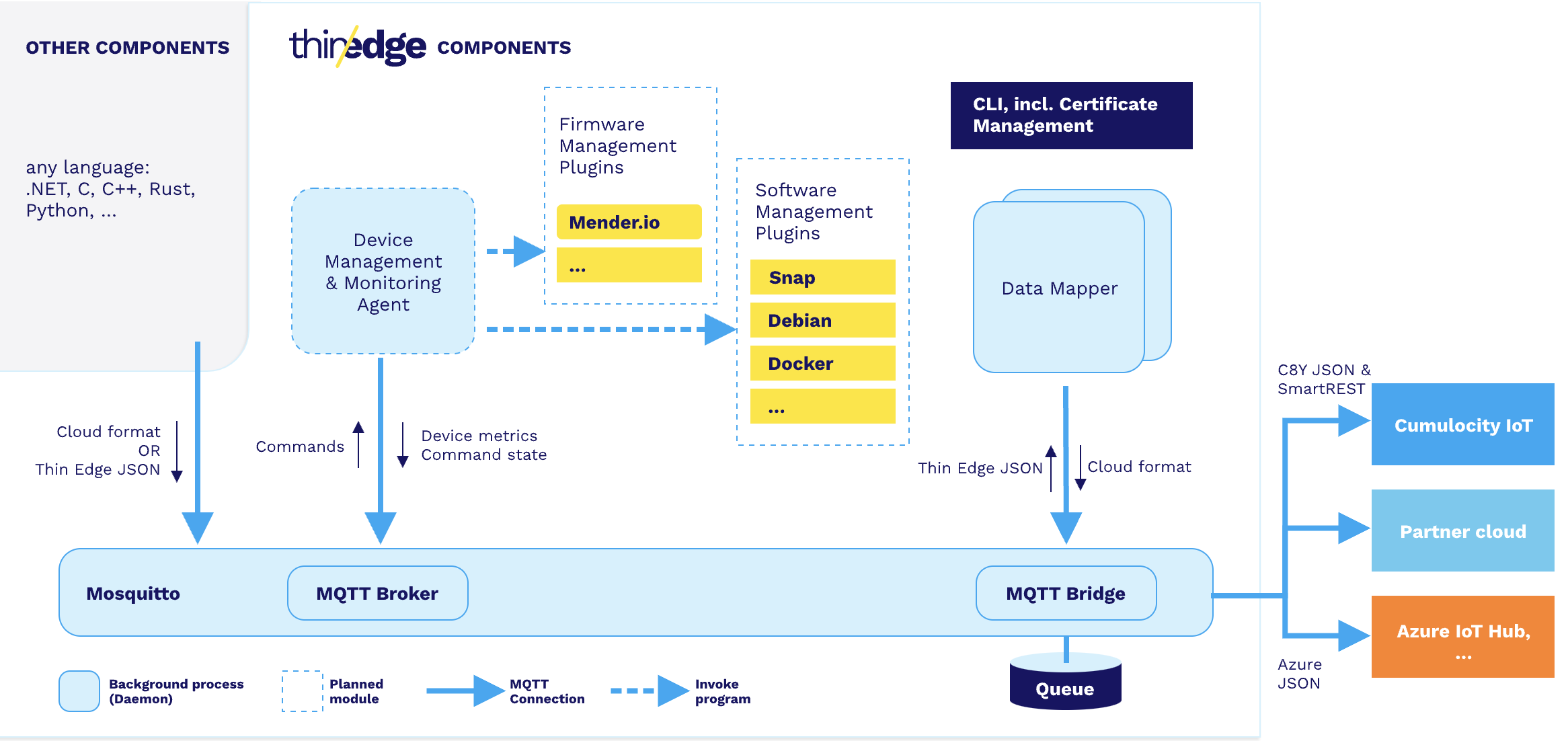}
    \caption{thin-edge.io Architecture. Source: https://thin-edge.io/}
    \label{fig:thin-edge}
\end{figure}

Thin-edge is open-source software that seamlessly integrates with platforms such as Cumulocity IoT, AWS IoT Core, Azure IoT Hub, and others. This integration enhances any existing edge server, providing robust capabilities for managing software and models throughout their deployment lifecycle, including installation, maintenance, and continuous performance monitoring. It complements Cumulocity IoT by serving as a specifically designed framework for edge computing environments. Thin-edge’s architecture (as shown in figure \ref{fig:thin-edge}) is tailored to enhance the performance and manageability of devices operating at the network’s edge, where responsiveness and local data processing are paramount. In thin-edge, the \textit{software} tab plays a crucial role by facilitating the distribution of software artifacts to each device, seamlessly integrating with Cumulocity IoT's repository to ensure that all edge devices are up-to-date and functioning optimally. Additionally, the \textit{services} tab within thin-edge provides a detailed overview of the metrics and performance for each software artifact, allowing for real-time monitoring and management of applications running on the edge.

Together, Cumulocity IoT and thin-edge create a synergistic ecosystem that is highly suited for deploying and managing ML solutions on edge devices. The built-in functionalities of both platforms provide a robust foundation for associating machine learning models and engines with edge devices, enabling sophisticated data processing and inferencing capabilities directly where data is collected.

\begin{figure}
    \centering
    \includegraphics[width=0.8\linewidth]{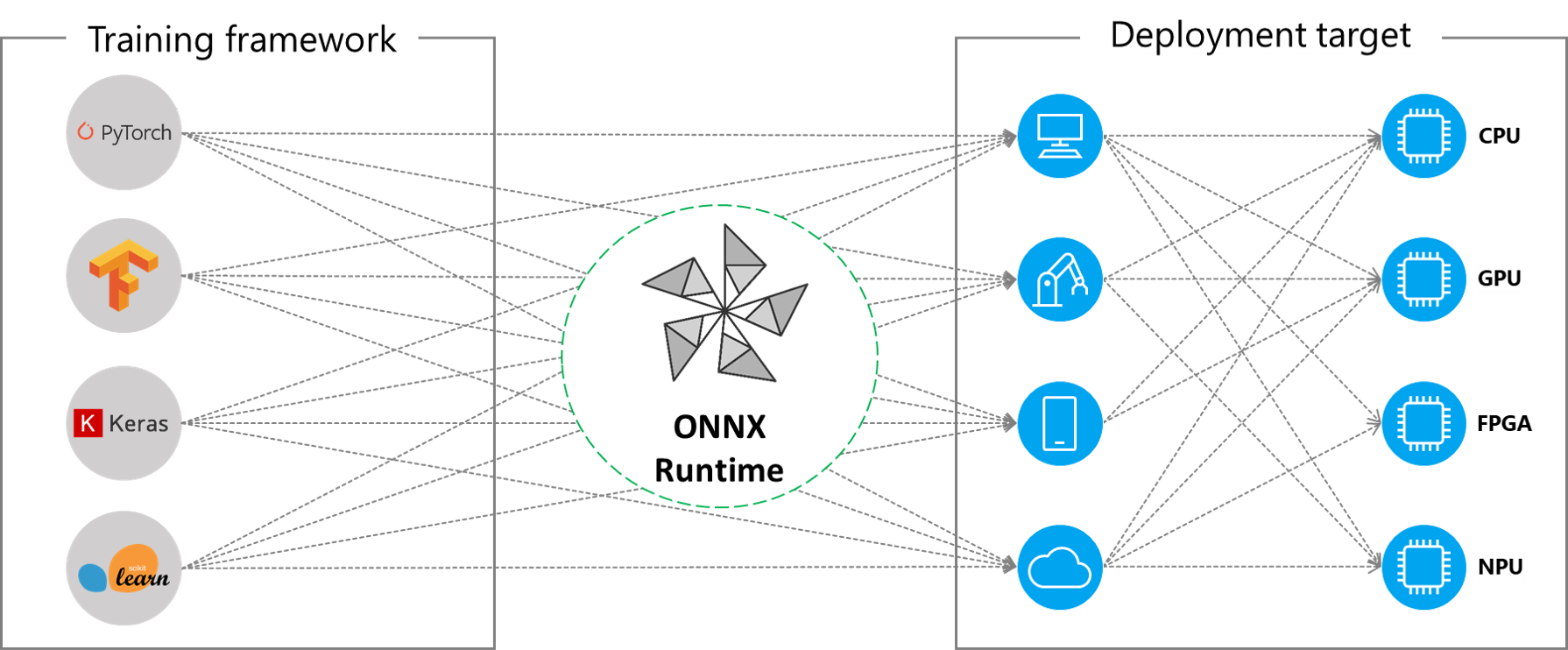}
    \caption{Overview of ONNX Runtime. Source: https://onnxruntime.ai/}
    \label{fig:onnxruntime}
\end{figure}

The ONNX (Open Neural Network Exchange) standard \footnote{\url{https://onnxruntime.ai/}} is an open-source initiative that enhances interoperability among AI models, allowing seamless transitions between training frameworks and deployment environments. It enables users to train models using popular frameworks such as TensorFlow, PyTorch, and scikit-learn, and then convert these models into the ONNX format for deployment using ONNX Runtime (as shown in figure \ref{fig:onnxruntime}). This facilitates a smoother workflow and ensures that models developed in various environments can be easily integrated and utilized across diverse platforms.

ONNX is designed to be highly versatile, supporting multiple deployment targets. It is capable of running on a wide range of hardware, from CPUs and GPUs to more specialized accelerators like FPGAs. Additionally, ONNX maintains compatibility across major operating systems including Windows, Linux, and macOS, ensuring that ONNX models can be deployed in virtually any computing environment.

Further enhancing its utility, ONNX supports model quantization which compresses model sizes and optimizes their performance. This is particularly beneficial for deployment on resource-constrained devices, as it not only speeds up inference times but also significantly reduces memory usage, making high-performance computing accessible on smaller, less powerful devices. Moreover, ONNX optimizes the computational graphs of models through techniques such as kernel fusion and constant folding. These optimizations are automatically applied to improve the efficiency and speed of model inference, ensuring optimal performance across different hardware platforms. Together, these features make ONNX a robust, flexible standard that is indispensable for developers looking to streamline and scale their machine learning operations.

%
\section{Architecture of \PaperAcronym{}}\label{sec:architecture}

This section describes an end-to-end Edge AI inferencing workflow utilizing Cumulocity IoT, thin-edge, and the ONNX standard. The following outlines each step of the process, from model creation and management to deployment and real-time inferencing at the edge. It highlights the seamless integration of these technologies to enhance IoT operations.

\begin{figure}
    \centering
    \includegraphics[width=1\linewidth]{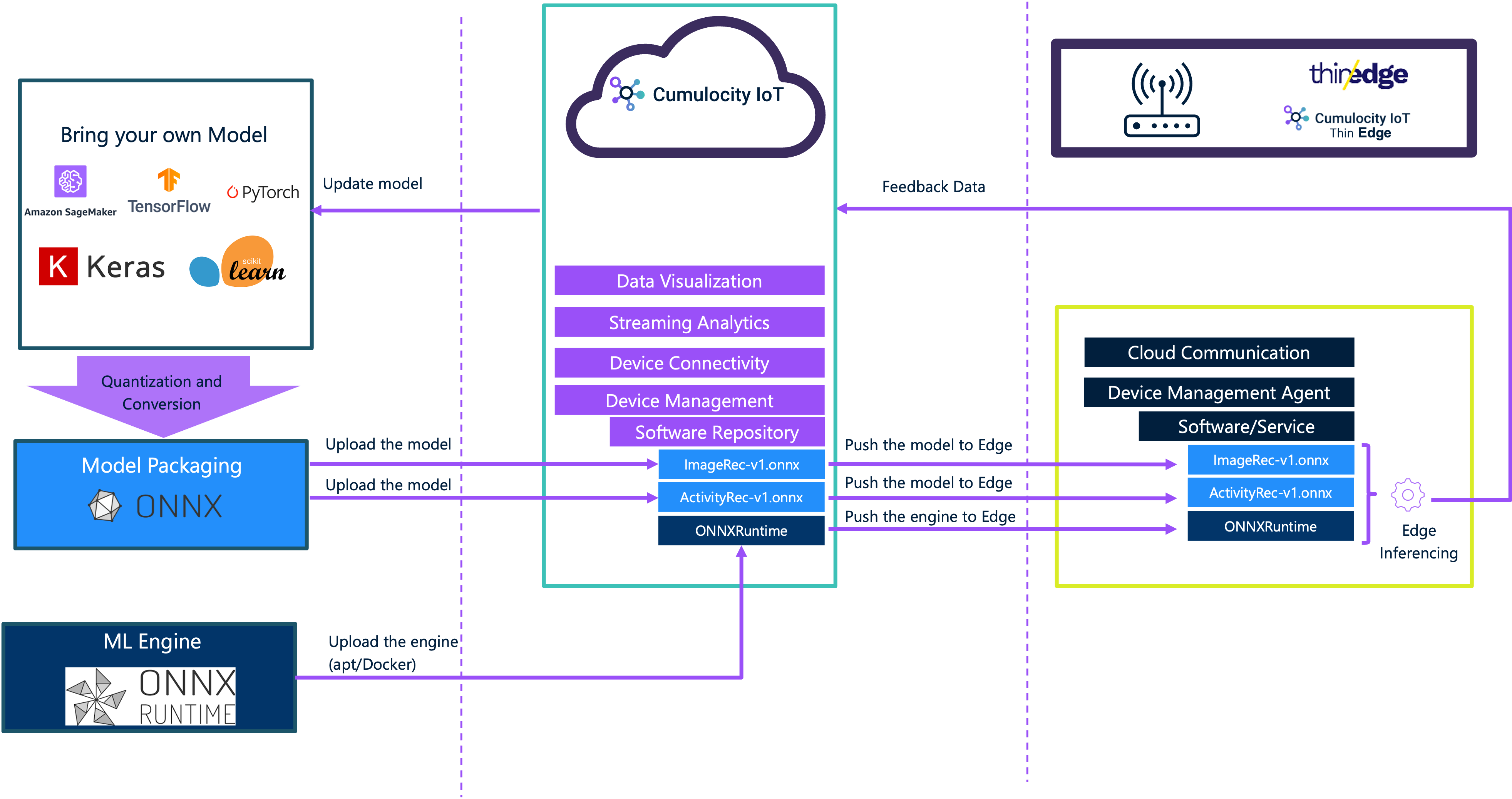}
    \caption{Edge AI inferencing workflow}
    \label{fig:EdgeAI-Architecture-1}
\end{figure}

As shown in figure \ref{fig:EdgeAI-Architecture-1}, in the left pane of our architecture, we begin with Model Creation. It allows users to bring their own AI/ML models, developed using advanced frameworks such as TensorFlow or Amazon SageMaker. These models undergo a quantization process to reduce their computational demands, making them more efficient and suitable for edge deployments. After optimization, the models are converted into the universally compatible ONNX format. Alongside the models, an ONNX Runtime is also prepared within a Docker container, ensuring that the runtime environment is portable and consistent across various deployment scenarios.

The middle pane focuses on the Cumulocity IoT Cloud Platform, which serves as the operational nucleus of the architecture. It features a Software Repository component where both ONNX models and their corresponding Docker containers housing the ONNX Runtime are uploaded and stored. This centralized repository facilitates the efficient management and deployment of these resources, allowing seamless distribution across a network of connected devices. Moving to the right pane, we focus on the Thin-edge component, which is crucial for device management at the edge. In this segment of our architecture, the ONNX models and the ONNX Runtime engine, sourced from the Cumulocity software repository, are delivered to individual edge devices. Operators have the ability to use the device console to deploy, manage, and fine-tune these models and their runtime environments. This process enables sophisticated edge inferencing capabilities directly on the devices, allowing them to perform complex analytics locally.

To enhance the functionality and versatility of the inferencing process, Python scripts can be added to the devices. These scripts are responsible for handling the essential steps of pre-processing, inferencing, and post-processing within the inferencing pipeline. This addition not only automates the process but also ensures that data is aptly prepared and analyzed, leading to more accurate and efficient outcomes. Furthermore, thin-edge also supports a Node-RED plugin \footnote{\url{https://github.com/thin-edge/tedge-nodered-plugin}}, offering users a visual interface to create and manage these workflows with ease. This feature allows for intuitive setup and maintenance of the data processing pipelines, making it accessible even to those with minimal coding expertise. This integration of Python scripts and Node-RED enhances the overall functionality of the edge devices, making them powerful tools for real-time data processing and decision-making in diverse IoT applications.

Lastly, the architecture includes a crucial feedback loop. Data and insights collected from Edge inferencing activities are transmitted back to the Cumulocity IoT Cloud. This data is not only visualized but also analyzed within the platform to assess model performance and operational efficiency. The insights obtained from this analysis are then utilized to refine and enhance the models. Improved versions are subsequently updated and redeployed to the edge devices via the same structured workflow, thus maintaining a continuous cycle of optimization and enhancement.

\begin{figure}
    \centering
    \includegraphics[width=1\linewidth]{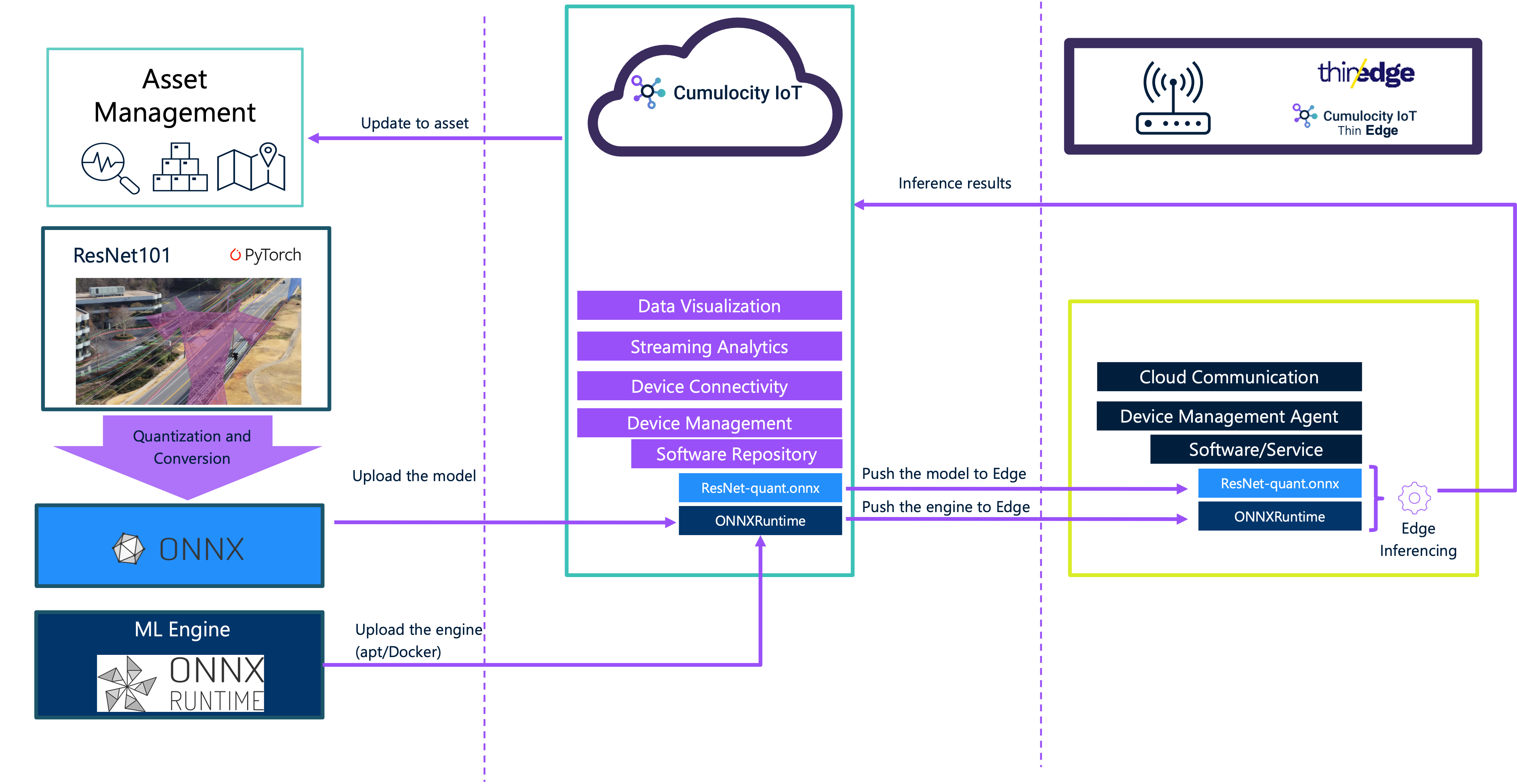}
    \caption{EdgeMLOps Framework for the Visual Quality Inspection use case}
    \label{fig:EdgeAI-Architecture-2}
\end{figure}

As shown in Figure \ref{fig:EdgeAI-Architecture-2}, we utilized the same architecture for our TTPLA dataset to solve the Visual Quality Inspection use case. The model was originally developed using PyTorch (c.f. Section \ref{sec:case_Study}). Once these models are trained, quantized, and converted to ONNX format, they integrate smoothly into our established workflow. Edge devices equipped with this technology, especially those with cameras, leverage these models for real-time image and video analysis. This capability to process and react to visual data directly at the edge showcases the practical application and efficiency of deploying high-performance models in edge environments. Such deployment facilitates immediate responses and decision-making based on visual inputs. This architecture not only supports but also enhances the functionality of IoT deployments across various sectors, presenting a versatile solution for a wide range of use cases.



%
\vspace{-3mm}\section{Quantization Benchmark}\label{sec:evaluation}
In this section, we present the results of our experiments performed to compare different quantization algorithms. In general, quantization reduces the precision of weights and activations of a model to reduce its size and resource consumption and accelerate its inference. The drawbacks include the potential loss of accuracy if the model is calibrated incorrectly, which can lead to hard-to-debug problems. Acceleration might require specialized hardware which might not be available on the existing target. 

To reduce the size of a floating point value, the value is scaled to either eight or lower-bit resolutions. To do that one can either scale symmetrically or asymmetrically. Symmetric quantization linearly reduces the range of the floating point to 8 bits. Asymmetric quantization calculates a custom zero point and thereby shifts the range of possible values. 
A quantize and corresponding de-quantize step replaces the original element and maintains its input and output shapes. Thereby the caller interaction does not change. 

\begin{figure*}[htbp]
	\centering
	\subfloat[Averge inference time]{\label{fig:average}\includegraphics[width=0.5\linewidth]{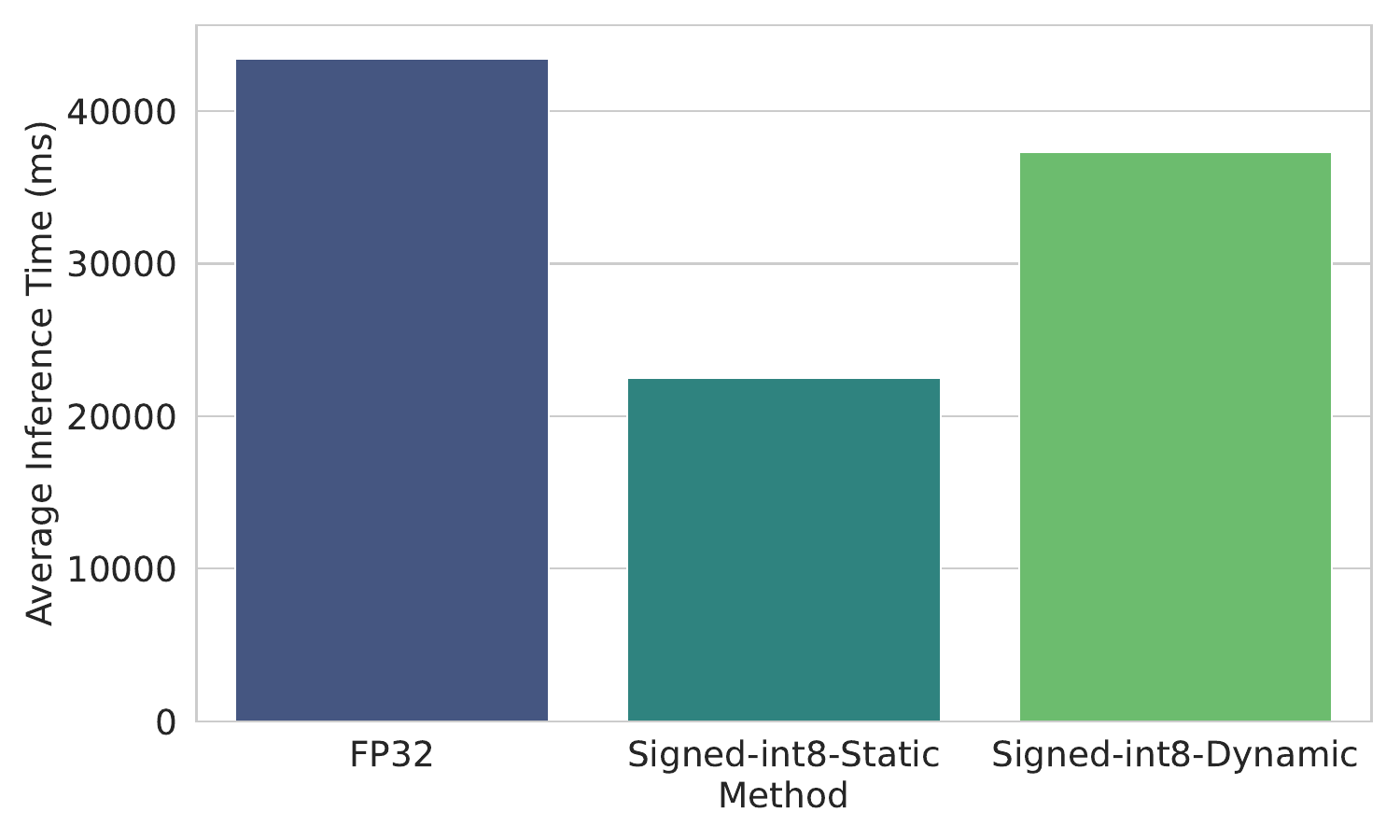}} 
	\subfloat[Distribution over various experiments]{\label{fig:boxplot}\includegraphics[width=0.5\linewidth]{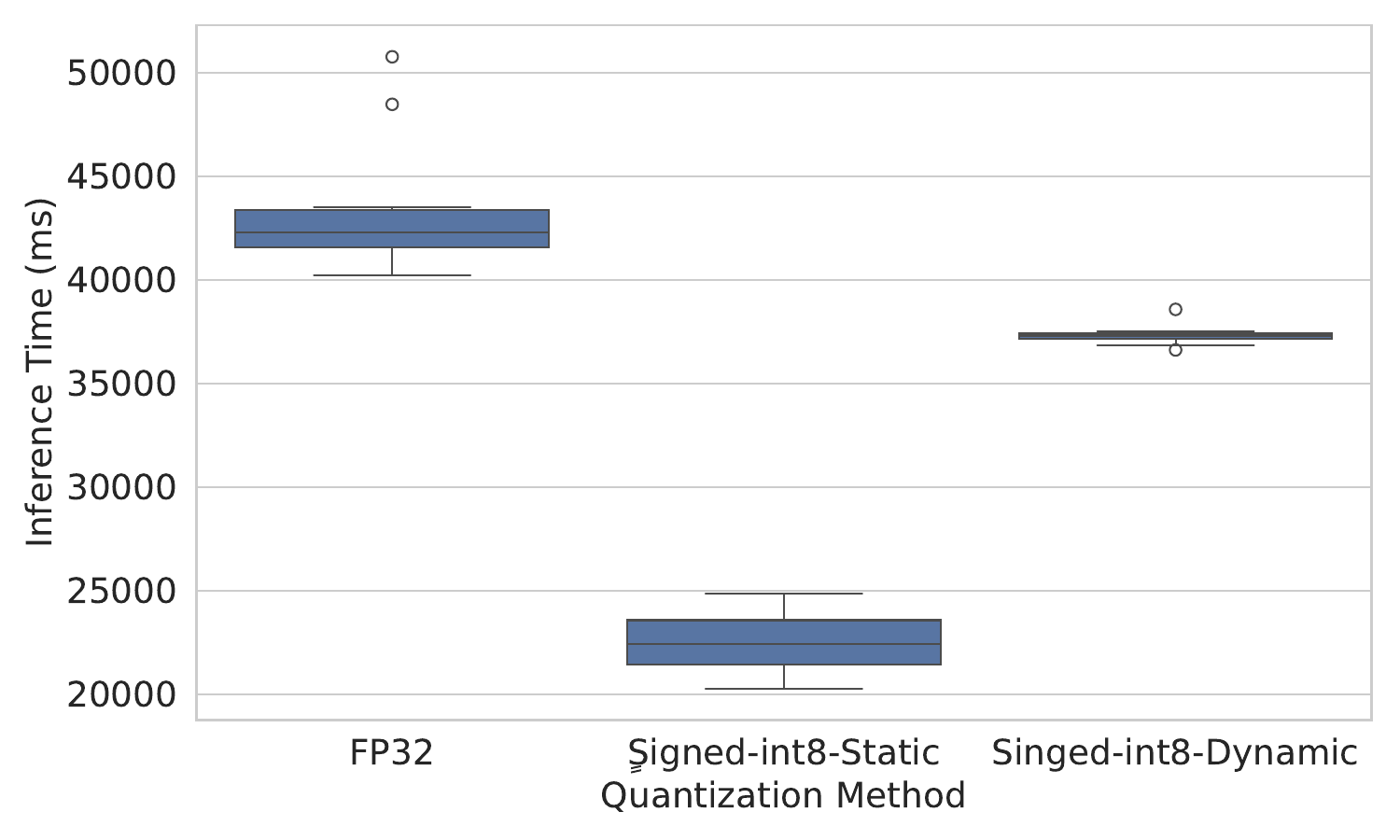}} 
    \caption{Quantization speed improvements on a Raspberry Pi 4 4GB}
	\label{fig:results} 
\end{figure*}

ONNX supports static and dynamic quantization but recommends the type based on the model and use case. The static calculation is favorable while using a well-known data distribution in development and inference, as dynamic quantization can be advantageous if data is not well-known beforehand, but is generally slower than its counterpart. ONNX recommends preprocessing ONNX models before quantization as compute graph optimizations can complicate debugging. Further, model validation can be done similarly to the original as input and output shapes remain identical. 
%
%
%
 Figure~\ref{fig:average} shows the average inference time for three different methods implemented on a Raspberry Pi 4: FP32, Signed-int8-Static, and Signed-int8-Dynamic. The FP32 method shows the highest inference time, indicating it requires the most computational resources. In contrast, the Signed-int8-Static method significantly reduces the inference time, demonstrating its efficiency in processing. The Signed-int8-Dynamic method offers a middle ground, with a moderate reduction in inference time compared to FP32. This comparison highlights the potential benefits of quantization for optimizing model performance on edge devices.

Figure~\ref{fig:boxplot} presents a box plot illustrating the inference time distribution across various experiments for three quantization methods: FP32, Signed-int8-Static, and Signed-int8-Dynamic. The FP32 method shows the highest and most consistent inference times, with minimal variability and a few outliers. In contrast, the Signed-int8-Static method significantly reduces inference time, displaying a wider range of variability but overall lower times. The Signed-int8-Dynamic method also reduces inference time compared to FP32, with a narrower distribution, indicating more consistent performance. This comparison highlights the efficiency gains achieved through quantization, particularly with the Signed-int8-Static method. The results obtained in these experiments suggest that the straightforward quantization with ONNX leads to a two-time speed improvement on a Raspberry Pi 4. After quantizing the model, we achieved the expected size reduction of approximately four. Both quantization methods used signed int8 values for the weights and showed small accuracy degradation.

\vspace{-3mm}\section{Conclusion}\label{sec:conclusion}
%
This work presented \PaperAcronym{}, a framework designed to streamline the deployment and management of ML models on edge devices using Cumulocity IoT and thin-edge.io. The VQI use case demonstrated the practical application of this framework, showcasing its ability to facilitate real-time asset condition monitoring. Our evaluation of quantization methods revealed substantial performance gains on a Raspberry Pi 4, with signed-int8 quantization achieving up to a two-fold reduction in inference time and a four-fold reduction in model size while maintaining acceptable accuracy. These findings underscore the effectiveness of \PaperAcronym{} in enabling efficient AI at the edge. Future work will explore advanced quantization techniques, federated learning strategies, and expanded application domains to further optimize and broaden the impact of \PaperAcronym{} in industrial IoT environments.

\bibliography{main}
\end{document}